\pdfoutput=1

\documentclass[11pt]{article}

\usepackage{acl}

\usepackage{times}
\usepackage{latexsym}

\usepackage[T1]{fontenc}

\usepackage[utf8]{inputenc}

\usepackage{microtype}

\usepackage{amsfonts,amssymb}
\usepackage{amsmath}
\usepackage{url}
\usepackage{booktabs}
\usepackage{multirow}
\usepackage{multicol}
\usepackage{makecell}

\usepackage{tikz}
\usepackage{subfigure}
\usepackage{pgfplots}
\usepackage{rotating,diagbox}
\usepackage{verbatim}
\usepackage{colortbl}

\usetikzlibrary{plotmarks}
\usetikzlibrary{patterns}
\usetikzlibrary{pgfplots.groupplots}
\usetikzlibrary{arrows}
\usetikzlibrary{decorations}
\usetikzlibrary{decorations.pathreplacing}
\usetikzlibrary{backgrounds}
\usetikzlibrary{fit}
\usetikzlibrary{positioning}
\usetikzlibrary{calc}
\usetikzlibrary{shadows}
\usetikzlibrary{shapes.multipart}

\definecolor{ugreen}{cmyk}{1,0,1,0.498}
\definecolor{lyyblue}{cmyk}{0.8278,0.3333,0,0.2941}
\definecolor{lyygreen}{cmyk}{0.6813,0,0.725,0.3725}
\definecolor{lyyred}{cmyk}{0,0.8855,0.8767,0.1098}
\definecolor{dblue}{cmyk}{1,0.5487,0,0.5569}

\title{When Structured Pruning meets Multilingual Pre-trained Language Models}
\title{Structured Pruning on Multilingual Pre-trained Language Models: \\ A Case Study}
\title{Exploring Pruning on Multilingual Pre-trained Language Models}
\title{Exploring Strategies for Pruning Multilingual Pre-trained \\ Language Models}

\title{Probing Structured Pruning on Multilingual Pre-trained Models: \\ Settings, Algorithms, and Efficiency}

\author{Yanyang Li$^1$\thanks{\ \ Collaborated work while doing an Alibaba DAMO Academy internship.}\ , Fuli Luo$^2$, Runxin Xu$^3$, Songfang Huang$^2$, Fei Huang$^2$, Liwei Wang$^1$ \\
$^1$Department of Computer Science and Engineering, The Chinese University of Hong Kong \\
$^2$Alibaba Group\\
$^3$Key Laboratory of Computational Linguistics, Peking University, MOE, China\\
\texttt{\{yyli21,lwwang\}@cse.cuhk.edu.hk, runxinxu@gmail.com} \\
\texttt{\{lfl259702,songfang.hsf,f.huang\}@alibaba-inc.com}\\
}

\begin{document}
\maketitle
\begin{abstract}





Structured pruning has been extensively studied on monolingual pre-trained language models and is yet to be fully evaluated on their multilingual counterparts.
This work investigates three aspects of structured pruning on multilingual pre-trained language models: settings, algorithms, and efficiency.
Experiments on nine downstream tasks show several counter-intuitive phenomena:
for settings, individually pruning for each language does not induce a better result;
for algorithms, the simplest method performs the best;
for efficiency, a fast model does not imply that it is also small.
To facilitate the comparison on all sparsity levels, we present \emph{Dynamic Sparsification}, a simple approach that allows training the model once and adapting to different model sizes at inference.
We hope this work fills the gap in the study of structured pruning on multilingual pre-trained models and sheds light on future research.
\end{abstract}

\section{Introduction}
\label{sec:intro}

Large-scale pre-trained monolingual language models like BERT \cite{devlin2019bert} and RoBERTa \cite{liu2019roberta} have shown promising results in various NLP tasks while suffering from their large model size and high latency.
Structured pruning has proven to be an effective approach to compressing and accelerating these large monolingual language models \cite{michel2019sixteen,wang-etal-2020-structured,prasanna-etal-2020-bert,liang-etal-2021-super}, making them practical for real-world applications.

Similarly, multilingual pre-trained models \cite{conneau2019cross,conneau-etal-2020-unsupervised,xue-etal-2021-mt5,luo-etal-2021-veco} are also powerful and even have more parameters. However, little attention has been paid to evaluating the effectiveness of structured pruning on these multilingual models.
Applying pruning to multilingual pre-trained models is non-trivial, as it typically involves many languages and needs to carefully design the roles of modules within the network. For example, most attention heads have little impact on the performance of monolingual pre-trained models \cite{michel2019sixteen,voita-etal-2019-analyzing}, while it is the opposite for multilingual pre-trained models (See Section \ref{sub-sec:analysis1} and also \citet{DBLP:journals/corr/abs-2109-12683}).

This work intends to examine how structured pruning reacts to multilingual pre-trained models.
We take the most representative multilingual pre-trained model family, XLM-R \cite{conneau-etal-2020-unsupervised,goyal2021larger} for our case study and evaluate the pruning performance on nine cross-lingual understanding tasks in XTREME \cite{hu2020xtreme}.
We investigate three aspects of structured pruning: settings, algorithms, and efficiency.

\noindent\textbf{Settings}
Traditional pruning produces a single small model, which is shared across languages (\texttt{shared} setting).
Recent work on multilingual translation \cite{li2020deep,lin-etal-2021-learning,xie-etal-2021-importance,gong2021adaptive} suggests that tailoring pruning to one language could achieve better results (\texttt{non-shared} setting).
However, our comprehensive experiments show that neither of the two settings can consistently outperform the other one (See Section \ref{sub-sec:result1}).

\noindent\textbf{Algorithms}
There exists a broad spectrum of pruning algorithms \cite{hoefler2021sparsity}, and it is impossible to test all of them considering the cost of pre-training.
We focus on two pruning algorithms that have been studied the most in monolingual pre-trained models: the regularization-based pruning \cite{louizos2018learning,wang-etal-2020-structured} (and our improved version) and the gradient-based pruning \cite{michel2019sixteen,prasanna-etal-2020-bert,liang-etal-2021-super} (See Section \ref{sec:pruning}).
We experimentally find that the simplest gradient-based pruning is more effective for XLM-R (See Section \ref{sub-sec:result1}).

\noindent\textbf{Efficiency}
One meaningful way to measure pruning algorithms is to study how the performance and speed of the pruned model vary with the sparsity \cite{hoefler2021sparsity}.
However, most pruning algorithms, including those we study in this work, require training the model for each specific sparsity.
This limitation makes comparisons against a range of sparsity levels infeasible due to the prohibitive training cost.
To solve this issue, we propose the \emph{Dynamic Sparsification} (DS for short), a simple method that parameterizes subnetworks at any sparsity level and shares their weights afterward (See Section \ref{sub-sec:dyna}).
DS only trains the model once but can obtain models at any sparsity level during inference.
Experiments on XNLI \cite{conneau-etal-2018-xnli} show that DS does not degrade the performance much while dramatically reducing the training cost.
Interestingly, we observe that the model size and inference speed are not strongly correlated in XLM-R. This observation suggests that one could not obtain a fast model by simply making the model small by using vanilla pruning algorithms (See Section \ref{sub-sec:result2}).

\section{Related Work}


%

\paragraph{Settings}
Recent multilingual translation research suggests that adapting subnetworks for each language or language pair rather than for all of them gives better results. Among them, \citet{li2020deep} train a shared multilingual model, then select layers for each language pair.
\citet{lin-etal-2021-learning} also prune a shared multilingual model for each language pair, though on the level of entries in weight matrices.
Instead, \citet{gong2021adaptive} prune attention heads and feedforward networks for each language.
\citet{xie-etal-2021-importance} first identify general and language-specific neurons in a shared multilingual network, then tune those neurons using the data of their corresponding language only.
These findings inspire us to extend from multilingual translation to see how \texttt{non-shared} pruning settings work on multilingual pre-training.

\paragraph{Algorithms}
There are many structured pruning techniques proposed for monolingual pre-trained language models recently.
\citet{michel2019sixteen} propose a simple gradient-based importance score to prune attention heads.
\citet{prasanna-etal-2020-bert,liang-etal-2021-super} extend to prune other components like the feedforward network of the Transformer \cite{vaswani2017attention}.
\citet{wang-etal-2020-structured} decompose the pre-trained model weights and apply $L_0$ regularization \cite{louizos2018learning} to regulate the ranks of decomposed weights.
\citet{sajjad2020effect} study layer pruning and show that directly dropping the top layers performs the best in fine-tuning.
\citet{peer2021greedy} further show that by carefully choosing layers to drop, structured pruning can achieve a performance close to those trained by knowledge distillation \cite{hinton2015distilling}.

\paragraph{Efficiency}
The pruning algorithms mentioned above need to train one network for each sparsity level used at inference.
\citet{hou2020dynabert} propose a dynamic structured pruning method based on \citet{michel2019sixteen}, which allows training the model once and making the inference with any size of the model.
Compared with our Dynamic Sparsification, \citet{hou2020dynabert}'s method cannot be applied to the \texttt{non-shared} setting as it needs to rearrange the network, i.e., producing a new model, for each language.
Cascading methods \cite{DBLP:conf/acl/SchwartzSSDS20,DBLP:conf/acl/XinTLYL20} can even adapt the network size for each instance.
Since cascading methods cannot perform batch inference and are only available for sentence classification tasks, we do not consider them in this work.


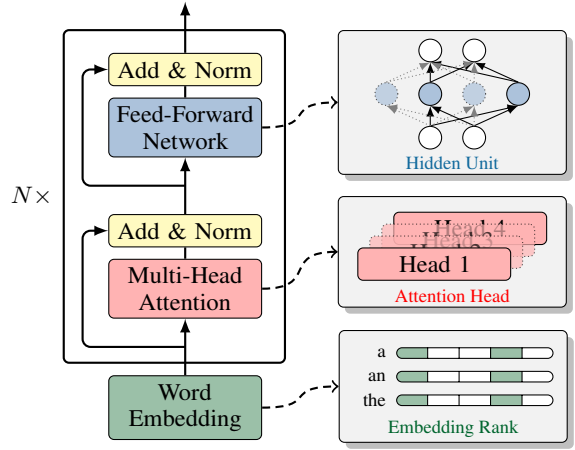
\begin{figure}[t!]
    \centering
    \begin{tikzpicture}[scale=1.5]
        \tikzstyle{wordnode} = [font=\scriptsize,align=center,inner sep=2pt]
        
        \begin{scope}[local bounding box=ENCODER]
            \tikzstyle{layernode} = [rectangle,draw,rounded corners=2pt,minimum height=0.2cm,minimum width=1.8cm,text width=1.8cm,inner sep=3pt,align=center,font=\small]
            
            \node[layernode,fill=ugreen!30!white,anchor=south] (embed) at (0,0) {Word\\Embedding};
            
            \node[layernode,fill=red!30!white,anchor=south] (attn) at ([yshift=0.5cm]embed.north) {Multi-Head\\Attention};
            \node[layernode,fill=yellow!30!white,anchor=south] (norm1) at ([yshift=0.1cm]attn.north) {Add \& Norm};
            
            \node[layernode,fill=lyyblue!30!white,anchor=south] (ffn) at ([yshift=0.5cm]norm1.north) {Feed-Forward\\Network};
            \node[layernode,fill=yellow!30!white,anchor=south] (norm2) at ([yshift=0.1cm]ffn.north) {Add \& Norm};
            
            \draw[-latex,thick] (embed) to (attn);
            \draw[thick] (attn) to (norm1);
            \draw[-latex,thick] (norm1) to (ffn);
            \draw[thick] (ffn) to (norm2);
            \draw[-latex,thick] (norm2) to ([yshift=0.6cm]norm2);
            
            \draw[-latex,thick,rounded corners=3pt] ([yshift=-0.25cm]attn.south) to +(-0.9cm,0) |- (norm1.west);
            \draw[-latex,thick,rounded corners=3pt] ([yshift=-0.25cm]ffn.south) to +(-0.9cm,0) |- (norm2.west);

            \begin{pgfonlayer}{background}
                \node[] (left) at ([shift={(-0.1cm,-0.1cm)}]attn.south west) {};
                \node[rectangle,draw,thick,fill=white,rounded corners=2pt,inner sep=0.3cm,fit=(attn) (norm2) (left),label={[font=\small]left:$N\times$}] (encoder) {};
            \end{pgfonlayer}
        \end{scope}

        \begin{scope}[local bounding box=EMBEDDING]
            \tikzstyle{embednode} = [rectangle,draw,rounded corners=2pt,inner sep=0pt,rectangle split,rectangle split horizontal,rectangle split every empty part={inner sep=0pt},rectangle split empty part width=0.4cm,rectangle split empty part height=4pt,rectangle split parts=5,rectangle split part fill={ugreen!30!white,white,white,ugreen!30!white,white}]
            
            \coordinate (embedbottomleft) at ([shift={(1.175cm,0.3cm)}]embed.south east);
            
            \node[embednode,anchor=south west,label={[font=\scriptsize,name=w1]left:the}] (e1) at (embedbottomleft) {};
            \node[embednode,anchor=south west,label={[font=\scriptsize,name=w2]left:an}] (e2) at ([yshift=3pt]e1.north west) {};
            \node[embednode,anchor=south west,label={[font=\scriptsize,name=w3]left:a}] (e3) at ([yshift=3pt]e2.north west) {};

            \begin{pgfonlayer}{background}
                \node[anchor=south] (embedbottom) at ([yshift=-0.25cm]e1.south) {};
                \node[rectangle,draw,fill=gray!10!white,rounded corners=2pt,inner sep=3pt,minimum width=3cm,fit=(e1) (e2) (e3) (w1) (w2) (w3) (embedbottom),drop shadow] (embedding) {};
            \end{pgfonlayer}
            
            \node[wordnode,anchor=south] () at (embedding.south) {\color{ugreen}Embedding Rank};
        \end{scope}
        
        \draw[->,thick,densely dashed] (embed.east) .. controls +(0.5cm,0) and +(-0.5cm,0) .. (embedding.west);

        \begin{scope}[local bounding box=ATTENTION]
            \tikzstyle{headnode} = [rectangle,draw,rounded corners=2pt,minimum height=0.2cm,minimum width=1.8cm,fill=red!30!white,text width=1.8cm,inner sep=3pt,align=center,font=\small]
            
            \coordinate (attentionbottomleft) at ([shift={(1.15cm,0.65cm)}]attn.south east);
            
            \node[headnode,anchor=south west] (h4) at (attentionbottomleft) {Head 4};
            \node[headnode,anchor=south west,densely dotted,opacity=0.7] (h3) at ([shift={(-3pt,-3pt)}]h4.south west) {Head 3};
            \node[headnode,anchor=south west,densely dotted,opacity=0.7] (h2) at ([shift={(-3pt,-3pt)}]h3.south west) {Head 2};
            \node[headnode,anchor=south west] (h1) at ([shift={(-3pt,-3pt)}]h2.south west) {Head 1};

            \begin{pgfonlayer}{background}
                \node[anchor=south] (attentionbottom) at ([yshift=-0.1cm]h1.south) {};
                \node[rectangle,draw,fill=gray!10!white,rounded corners=2pt,inner sep=0.2cm,minimum width=3cm,fit=(h1) (h2) (h3) (h4) (attentionbottom),drop shadow] (attention) {};
            \end{pgfonlayer}
            
            \node[wordnode,anchor=south] () at (attention.south) {\color{red}Attention Head};
        \end{scope}
        
        \draw[->,thick,densely dashed] (attn.east) .. controls +(0.5cm,0) and +(-0.5cm,0) .. (attention.west);

        \begin{scope}[local bounding box=FFN]
            \tikzstyle{hiddennode} = [circle,draw,rounded corners=2pt,minimum size=0.3cm,inner sep=0pt,fill=white]
            
            \coordinate (ffnbottomleft) at ([shift={(1.4cm,0.1cm)}]ffn.south east);
            
            \node[hiddennode,anchor=south west] (i1) at (ffnbottomleft) {};
            \node[hiddennode,anchor=west] (i2) at ([xshift=5pt]i1.east) {};
            \node[hiddennode,anchor=south,fill=lyyblue!30!white] (n2) at ([yshift=5pt]i1.north) {};
            \node[hiddennode,anchor=south,fill=lyyblue!30!white,densely dotted,opacity=0.7] (n3) at ([yshift=5pt]i2.north) {};
            \node[hiddennode,anchor=east,fill=lyyblue!30!white,densely dotted,opacity=0.7] (n1) at ([xshift=-5pt]n2.west) {};
            \node[hiddennode,anchor=west,fill=lyyblue!30!white] (n4) at ([xshift=5pt]n3.east) {};
            \node[hiddennode,anchor=south] (o1) at ([yshift=5pt]n2.north) {};
            \node[hiddennode,anchor=south] (o2) at ([yshift=5pt]n3.north) {};
            
            \foreach \i in {2,4} {
                \foreach \j in {1,2} {
                    \draw[-latex,black] (i\j.north) to (n\i.south);
                    \draw[-latex,black] (n\i.north) to (o\j.south);
                }
            }
            
            \foreach \i in {1,3} {
                \foreach \j in {1,2} {
                    \draw[-latex,gray,densely dotted] (i\j.north) to (n\i.south);
                    \draw[-latex,gray,densely dotted] (n\i.north) to (o\j.south);
                }
            }

            \begin{pgfonlayer}{background}
                \node[anchor=south] (ffnbottom) at ([yshift=-0.15cm]i1.south) {};
                \node[rectangle,draw,fill=gray!10!white,rounded corners=2pt,inner sep=3pt,minimum width=3cm,fit=(i1) (i2) (n1) (n2) (n3) (n4) (o1) (o2) (ffnbottom),drop shadow] (feedforward) {};
            \end{pgfonlayer}
            
            \node[wordnode,anchor=south] () at (feedforward.south) {\color{lyyblue}Hidden Unit};
        \end{scope}
        
        \draw[->,thick,densely dashed] (ffn.east) .. controls +(0.5cm,0) and +(-0.5cm,0) .. (feedforward.west);
    \end{tikzpicture}
    \caption{The left is the Transformer encoder, the right is the components that will be pruned at each layer.}
    \label{fig:model}
\end{figure}



\section{Background}

In this section, we briefly review the structure of XLM-R \cite{conneau-etal-2020-unsupervised}, a Transformer encoder \cite{vaswani2017attention} pre-trained by masked language modeling task \cite{devlin2019bert}.
We also revisit how conventional structured pruning algorithms are applied to Transformers by introducing additional gating variables and setting appropriate values to them (See Figure \ref{fig:model} and also \citet{prasanna-etal-2020-bert,liang-etal-2021-super}).
The XLM-R model consists of $N$ layers.
Each layer is made of the multihead attention and feedforward networks, followed by the residual connection and layer normalization.


\paragraph{Attention}
Following \citet{michel2019sixteen}'s formula, the multihead attention is written as:
\begin{equation}
    \mathrm{MHA}(X)=\sum^H_{i=1} G_{h,i}\mathrm{head}_i
    \label{eqn:mha}
\end{equation}
where $H$ is the number of heads, $\mathrm{head}_i$ is the output of $i$-th head and $G_{h,i}$ is the $i$-th entry of the gating variables $G_h \in \mathbb{R}^H$.
$G_{h,i}$ indicates whether the head $i$ will be pruned.
$G_{h,i}$ is set to 1 to retain that head and 0 if to drop it.
Different pruning algorithms will have their own ways to determine the values of $G_{h}$. 

\paragraph{Feedforward Network}
The feedforward network contains two linear projections with GeLU activation \cite{hendrycks2016gaussian} in between:
\begin{equation}
    \mathrm{FFN}(X)=(\mathrm{GeLU}(XW_1+b_1)\odot G_f)W_2+b_2
    \label{eqn:ffn}
\end{equation}
where $W_1 \in \mathbb{R}^{d \times d_f}$, $b_1 \in \mathbb{R}^{d_f}$, $W_2 \in \mathbb{R}^{d_f \times d}$ and $b_2 \in \mathbb{R}^{d}$ are weights of the feedforward network and $d_f$ is the hidden size.
$\odot$ denotes the Hadamard product and $G_f \in \mathbb{R}^{d_f}$ is a gating vector with a value in the range of [0, 1].
$G_f$ functions similar to $G_h$ in multihead attention, except that $G_f$ controls the activation of hidden units.

\paragraph{Embedding}
To prune the large embedding matrix $E$ (occupying 69\% of all parameters), we decompose it via low-rank approximation as in \citet{lan2019albert}:
\begin{equation}
    E=\hat{E}\ \mathrm{diag}(G_e)P
\end{equation}
where $\hat{E} \in \mathbb{R}^{v \times d}$ and $P \in \mathbb{R}^{d \times d}$ are the decomposed matrices of $E$.
$v$ is the vocabulary size.
$G_e \in \mathbb{R}^{d}$, governing the rank of $E$, is a gating vector similar to $G_h$ and $G_f$.
$\mathrm{diag}(G_e)$ converts $G_e$ to a diagonal matrix.
The right part of Figure \ref{fig:model} is an illustration of the components (such as hidden units, attention heads, and embeddings) that will be pruned.

\section{Extending Pruning Algorithms to Pruning Settings}
\label{sec:pruning}

This section will first introduce pruning algorithms that we study and then describe how to adapt them to two pruning settings. The first is the \texttt{shared} setting that shares the pruned network across languages (default setting that all pruning algorithms could run on), and the second is the \texttt{non-shared} setting that prunes one subnetwork for each language \cite{xie-etal-2021-importance,gong2021adaptive}.

\subsection{Gradient-based Pruning} Gradient-based pruning \cite{michel2019sixteen} computes the importance score of each component, e.g., heads in Eq. \ref{eqn:mha}.
Then it sets the gating variable of a component, e.g., $G_{h,i}$ in Eq. \ref{eqn:mha}, to 1 if its importance score is larger than a threshold and 0 otherwise.
Taking an attention head $i$ as an example, its importance score is defined as:
\begin{equation}
    I_{\mathrm{head}_i}=\mathbb{E}_{X \sim \mathbf{X}}\left|\mathrm{head}^T_i\frac{\partial \mathcal{L}_{\mathrm{MLM}}(X)}{\partial \mathrm{head}_i}\right|
\end{equation}
where $\mathbf{X}$ is the data distribution and we choose the validation set as $\mathbf{X}$ in practice, $\mathcal{L}_{\mathrm{MLM}}$ is the masked language modeling loss \cite{devlin2019bert}.
The values of gating variables are set and frozen after pre-training.
An additional phase of pre-training is further employed to update network parameters to recover performance loss brought by pruning.

Extending gradient-based pruning to the \texttt{non-shared} setting is straightforward: to prune for one language, we use data of that language to compute a unique set of gating variables $G=\{G_h,G_f,G_e\}$ for it.

\subsection{Regularization-based Pruning}\label{sec:l0}
The $L_0$ norm has been widely used in many areas, including signal processing \cite{zhang2010analysis,xu2011image} to induce sparsity. In neural networks, regularization-based pruning, also referred to as $L_0$ regularization \cite{louizos2018learning}, defines a differentiable $L_0$ norm on the gating variables $G=\{G_h,G_f,G_e\}$. It controls the network sparsity by learning the values of $G$ during pre-training.
Taking a gating variable $g \in G$ as an example, it is modeled as:
\begin{eqnarray}
    \label{eqn:u}
    u&\sim&U(0,1)\\
    \label{eqn:s}
    s&=&\mathrm{sigmoid}((\log u/(1 - u)+ \alpha)/\beta) \\
    \hat{s}&=&s \times (r - l) + l\\
    \label{eqn:g}
    g&=&\min(1, \max(0, \hat{s}))
\end{eqnarray}
where $U$ is the uniform distribution, $l<0$ and $r>1$ are two fixed constants, $\beta$ is the temperature and $\alpha$ is a learnable parameter of $g$.
During training, $u$ is sampled for each $g$ separately.
At inference, Eq. \ref{eqn:s} becomes $s=\mathrm{sigmoid}(\alpha)$.
Compared with gradient-based pruning, the importance score in $L_0$ regularization is the learnt $\alpha$ and the threshold is fixed to $\mathrm{sigmoid}^{-1}\left(-\frac{l}{r-l}\right)$.

The $L_0$ regularization term of $g$ is:
\begin{equation}
    ||g||_0=\mathrm{sigmoid}\left(\alpha-\log(-l/r)\right)
\end{equation}
and the overall $L_0$ regularization term is\footnote{In practice we weigh the $L_0$ regularization term of gating variables (See Appendix \ref{app:weight}).}:
\begin{equation}
    \mathcal{L}_{L_0}=||G||_0=\sum_{g \in G}||g||_0
    \label{eqn:l0}
\end{equation}
$\mathcal{L}_{L_0}$ will be multiplied by a hyper-parameter $\lambda_1$ and added to the pre-training loss $\mathcal{L}_{\mathrm{MLM}}$.



\subsubsection{Improved $L_0$ Regularization} \label{sub-sec:improved-l0}

Two issues of the previous native $L_0$ regularization emerge in practice:
1) The hyper-parameter $\lambda_1$ does not relate to the model sparsity. It requires several expensive try-outs training runs to find an appropriate setup that can reach desired sparsity \cite{wang-etal-2020-structured}.
2) If we extend $L_0$ regularization to \texttt{non-shared} setting as done in gradient-based pruning, it easily converges to an optimum where every language shares the network \cite{gong2021adaptive}.
This falls back to the \texttt{shared} setting.
Thus, we propose two corresponding solutions as below:

\paragraph{1) Sparsity Constraint} To address the first issue, we add a sparsity constraint to Eq. \ref{eqn:l0}:
\begin{equation}
    \mathcal{L}_{L_0}=\sum^l_{i=1} \left|||G^i||_0 - t\right|
    \label{eqn:constraint}
\end{equation}
where $l$ is the number of languages and $G^i$ denotes the set of gating variables for language $i$.
This loss term will keep the subnetwork size of each language close to the targeted size $t$.\footnote{Adding a Lagrange multiplier \cite{wang-etal-2020-structured} is also doable, but we find this simple $L_1$-like loss is similarly effective and easy to implement.}

\paragraph{2) Diverse Subnetwork} To address the second issue, we introduce a diversity loss term to encourage the model to find a distinct subnetwork for each language.
It is achieved by diagonalizing the gram matrix of gating variables $\bar{G}=[G^1;\cdots;G^l]$:
\begin{equation}
    \mathcal{L}_{\mathrm{diag}}=||P \odot \bar{G}\bar{G}^T \odot (\mathbf{1}-\mathbf{I})||_1
    \label{eqn:diag}
\end{equation}
where $\mathbf{1}$ is a matrix of ones and $\mathbf{I}$ is the identity matrix.
$P \in \mathbb{R}^{l \times l}$ is used to introduce linguistic prior and is a matrix of ones by default.

Eq. \ref{eqn:diag} will penalize each language pair equally.
Intuitively, the subnetworks of two languages that are close, e.g., English and Spanish, should not be penalized.
Thus we add linguistic prior $P_{ij}=0$ when the $i$-th and $j$-th languages belong to the same language family (See Appendix \ref{app:family}) and 1 otherwise.

To the end, the loss $\mathcal{L}$ we used in pre-training is:
\begin{equation}
    \mathcal{L}=\mathcal{L}_{\mathrm{MLM}} + \lambda_1 \mathcal{L}_{L_0} + \lambda_2 \mathcal{L}_{\mathrm{diag}}
\end{equation}

Note that the parameter of the gating variable $\alpha$ is randomly initialized.
We find that tuning only $\alpha$ in the first few epochs is crucial to obtain better performance.
If no further notice, we will use this improved $L_0$ regularization for experiments with \texttt{non-shared} setting and the native $L_0$ regularization for \texttt{shared} setting.

\section{Empirical Study of Algorithms and Settings for Multilingual Pruning}

\subsection{Experimental Setup}
\label{sec:setup}

\begin{table*}[t!]
    \centering
    \setlength{\tabcolsep}{4pt}
    {\small
    \begin{tabular}{lrcccccccccc}
        \toprule
        Task & \multirow{3}*{Sparsity} & XNLI & PAWS-X & POS & NER & XQuAD & MLQA & TyDiQA & BUCC & Tatoeba & \multirow{3}*{\underline{\textbf{Avg}}} \\
        Metrics & & Acc. & Acc. & F1 & F1 & F1/EM & F1/EM & F1/EM & F1 & Acc. & \\
        \#Languages & & 15 & 7 & 33 & 40 & 11 & 7 & 9 & 5 & 33 & \\
        \midrule
        \multicolumn{12}{l}{\emph{\textbf{Cross-lingual Transfer}: Fine-tune model on English training set and test on all languages.}}\\
        XLM-R & 0\% & 74.8 & 85.4 & 74.0 & 61.9 & 69.2/53.0 & 59.9/44.3 & 51.3/32.4 & 63.3 & 53.4 & 60.2 \\
        DistilBERT & 50\% & 70.3 & 82.9 & 72.1 & 56.1 & 60.5/44.3 & 52.4/37.4 & 39.4/23.0 & 44.2 & 45.3 & 52.3 \\
        $L_0$ (non-shared) & 50\% & 68.6 & 83.3 & 68.3 & 53.4 & 59.8/43.2 & 49.6/34.6 & 35.2/19.8 & 52.5 & 43.8 & 51.0 \\
        $L_0$ (shared) & 20\% & 65.3 & 80.9 & 68.4 & 52.0 & 54.8/38.7 & 45.7/30.7 & 26.8/13.5 & 34.2 & 41.1 & 46.0 \\
        Grad (non-shared) & 50\% & 68.6 & 83.9 & 68.3 & 53.9 & 60.6/44.2 & 52.3/36.7 & 40.5/22.6 & \textbf{57.5} & \textbf{48.6} & 53.1 \\
        Grad (shared) & 50\% & \textbf{70.4} & \textbf{84.7} & \textbf{72.4} & \textbf{57.4} & \textbf{64.2/48.3} & \textbf{56.1/40.5} & \textbf{45.2/28.0} & 46.6 & 40.5 & \underline{\textbf{54.5}} \\
        \midrule
        \multicolumn{12}{l}{\emph{\textbf{Translate-Train-All}: Fine-tune model on English training data and translated data of other languages.}}\\
        XLM-R & 0\% & 79.1 & 89.2 & 89.5 & 88.0 & 72.7/58.2 & 58.2/42.8 & 72.1/57.5 & - & - & 70.7 \\
        DistilBERT & 50\% & 75.8 & 87.3 & \textbf{88.9} & 87.1 & 69.0/54.3 & 55.0/39.6 & 68.6/53.7 & - & - & 67.9 \\
        $L_0$ (non-shared) & 50\% & 76.3 & 87.8 & 87.9 & 86.8 & 69.3/54.2 & 54.7/39.2 & 67.8/52.5 & - & - & 67.7 \\
        $L_0$ (shared) & 20\% & 73.4 & 86.0 & 87.5 & 85.1 & 65.1/50.1 & 51.2/35.6 & 61.2/45.9 & - & - & 64.1 \\
        Grad (non-shared) & 50\% & 76.6 & 88.2 & 87.3 & 86.6 & 68.9/53.6 & 55.2/39.5 & 68.6/53.7 & - & - & 67.8 \\
        Grad (shared) & 50\% & \textbf{76.8} & \textbf{88.4} & 88.4 & \textbf{88.0} & \textbf{70.1/55.0} & \textbf{56.7/40.7} & \textbf{69.5/54.6} & - & - & \underline{\textbf{68.8}} \\
        \bottomrule
    \end{tabular}
    }
    \caption{XTREME results (Sparsity is the portion of dropped parameters in the Transformer encoder, and thus higher sparsity denotes smaller size.). We compare one representative distillation method (denoted as \texttt{DistilBERT},  \citet{sanh2019distilbert}) and two representative structured pruning methods: gradient-based pruning (denoted as \texttt{Grad}) and regularization-based pruning (denoted as \texttt{$L_0$}), under two settings (described in Section \ref{sec:pruning}: \texttt{shared} and \texttt{non-shared}). \textbf{Bold} denotes the best results among 50\% sparsity.
    Note that since BUCC and Tatoeba do not have the translated training data, we do not report their translate-train-all results. }
    \label{tab:main}
\end{table*}

\paragraph{Pre-training} Our pruned models are trained on the CC-100 corpus \cite{wenzek-etal-2020-ccnet}.
We choose 100 languages with a total size of 2.2TB for training, which is consistent with those used in XLM-R \cite{conneau-etal-2020-unsupervised}.
The development set we used to induce the importance score for pruning is 3K randomly selected samples from the CC-100 corpus per language.

Our model is a 12-layer Transformer with a 768 embedding size and a 3072 hidden size.
It is pruned and continually trained based on the publicly available XLM-R model for 150K steps with a batch size of 2048 and a learning rate of 0.0002.
Other hyper-parameters remain the same as in the original paper  \cite{conneau-etal-2020-unsupervised}.
We train our model on 32 Nvidia Tesla V100 32GB GPUs with mixed-precision training.
It takes roughly 7-10 days to pre-train one model.
For inference, we use 1 Nvidia Tesla V100 32GB GPU and Intel(R) Xeon(R) Platinum 8269CY CPU @ 2.50GHz to estimate the GPU and CPU throughput (with a batch size of 128 for GPU and 1 for CPU).


\paragraph{Fine-tuning} We evaluate the pruned models on 9 downstream tasks from XTREME \cite{hu2020xtreme}.
These tasks can be classified into four different categories:
(1) sentence-pair classification: XNLI \cite{conneau-etal-2018-xnli}, PAWS-X \cite{yang-etal-2019-paws};
(2) structured prediction: POS \cite{d701bee1cabe492caf36340d6341e27b}, Wikiann NER \cite{pan-etal-2017-cross};
(3) question answering: XQuAD \cite{artetxe-etal-2020-cross}, MLQA \cite{lewis-etal-2020-mlqa}, TyDiQA \cite{clark-etal-2020-tydi};
(4) sentence retrieval: BUCC2018 \cite{zweigenbaum-etal-2017-overview}, Tatoeba \cite{TACL1742}.
The hyper-parameter setup of fine-tuning could be found in Appendix \ref{app:hyper}.

Following previous work \cite{hu2020xtreme}, we study the pruned models in two fine-tuning settings:
\emph{Cross-lingual Transfer} (a.k.a., zero-shot) and \emph{Translate-Train-All} (a.k.a., multi-task).
Note that for the two sequence labelling tasks POS and NER, translation cannot give us the correct training labels.
We thus use human-annotated data for translate-train-all training on them.


\subsection{Results}
\label{sub-sec:result1}

Table \ref{tab:main} shows the fine-tuning results of using different methods to prune XLM-R to 50\% sparsity (also the value of $t$ in Eq. \ref{eqn:constraint}).
We follow the convention of \citet{prasanna-etal-2020-bert} to compute the sparsity of the encoder, which excludes the embeddings in the calculation.
For \texttt{DistilBERT}, we remove half of the original layers of XLM-R as done in \citet{sanh2019distilbert}.
Note that in Table \ref{tab:main} (the rows of ``$L_0$ (shared)''), regularization-based pruning with \texttt{shared} setting has a lower sparsity (20\%).\footnote{We have tried various hyper-parameters settings to pre-train models toward 50\% sparsity (for a fair comparison with \texttt{DistilBERT}) using vanilla $L_0$, but the resulting sparsity is either too high ($\ge$70\%) or too low ($\le$20\%).
This is in line with the trainability issue of $L_0$ as indicated in Section \ref{sec:l0}.}

\paragraph{Gradient-based pruning performs better than regularization-based pruning.}
Table \ref{tab:main} shows that vanilla $L_0$ in \texttt{shared} setting has more parameters (20\% sparsity) but performs worse than gradient-based pruning with fewer parameters (50\% sparsity).
Despite that our proposed improved $L_0$ works better (\texttt{non-shared} setting), it still underperforms the gradient-based pruning counterpart.
This is because regularization-based pruning keeps modifying the subnetwork structure when weights are updating, which might introduce too much noise during training.
Gradient-based pruning, on the other hand, keeps the pruned network unchanged and adapts weights only.
Despite that some works \cite{hoefler2021sparsity} suggest that regularization-based pruning should be preferred, it might not be the same conclusion for XLM-R.

\begin{table}[t]
    \centering
    \setlength{\tabcolsep}{1pt}
    {\small
    \begin{tabular}{lcccccc}
        \toprule
        \makecell[c]{\textbf{Methods}} & Sparsity & XNLI & POS & NER & TyDiQA & \underline{\textbf{Avg}} \\
        \midrule
        $L_0$ & 20\% &  73.4 & 87.5 & 85.1 & 61.2/45.9 & 74.9 \\
        Impv. $L_0$ & 50\% & 76.3 & \textbf{87.9} & \textbf{86.8} & 67.8/52.5 & 77.8 \\
        Impv. $L_0$ + Distil  & 50\% & \textbf{76.4} & 87.5 & 86.7 & \textbf{69.5/54.6} & \underline{\textbf{78.2}} \\
        \bottomrule
    \end{tabular}
    }
    \caption{The results of the improved $L_0$ (Impv. $L_0$) regularization-based pruning (See Section~\ref{sub-sec:improved-l0}).}
    \label{tab:l0}
\end{table}

\paragraph{Neither of the pruning settings performs consistently better.}
Previous work on multilingual translation has suggested that \texttt{non-shared} setting provides consistent gains, as this way allows the pruned model to adapt for each language  \cite{li2020deep,lin-etal-2021-learning,xie-etal-2021-importance,gong2021adaptive}.
However, this is not the case for XLM-R.
As shown in Table \ref{tab:main}, regularization-based pruning ($L_0$) works the best with the \texttt{non-shared} settings\footnote{Non-shared model with more parameters dropped (50\% sparsity) is better than shared model with fewer parameters dropped (20\% sparsity).}, but for gradient-based pruning it is the \texttt{shared} setting.
We analyze that this is because XLM-R covers more low-resource languages (100 languages in XLM-R vs. 24 in most multilingual translation research), which makes sharing the subnetwork for a universal representation more preferable \cite{aharoni-etal-2019-massively}.


\paragraph{Simple distillation performs less effective than pruning.}
For most tasks, distillation is not as effective as pruning.\footnote{Although adopting advanced distillation techniques might improve the result, the pruning algorithm is also simple here.}
This might be that distillation prunes a whole layer, while more fine-grained components are pruned in structured pruning.
But combining distillation with pruning could provide some gain, as shown in Table \ref{tab:l0}.

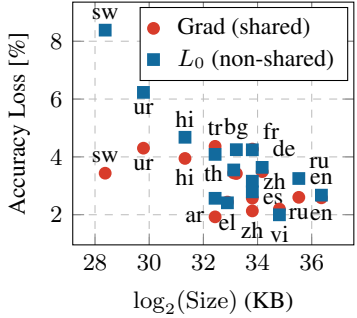
\begin{figure}[t!]
    \centering
    \begin{tikzpicture}
        \begin{axis}[
            width=0.33\textwidth,
            height=0.3\textwidth,
            legend cell align={left},
            enlargelimits=0.15,
            legend style={
                font=\small,
                column sep=5pt,
                legend columns=1,
            },
            yticklabel style={/pgf/number format/fixed,/pgf/number format/precision=1},
            ylabel={Accuracy Loss [\%]},
            ylabel near ticks,
            xlabel={$\log_2(\mathrm{Size})$ (KB)},
            xlabel near ticks,
            xmajorgrids=true,
            ymajorgrids=true,
            grid style=dashed,
            xtick={28,30,...,38},
            ytick={0,2,...,8},
            every tick label/.append style={font=\small},
            label style={font=\small},
            ylabel style={yshift=0pt},
        ]
            \addplot [
                lyyred,thick,mark=*,
                nodes near coords,
                scatter,
                every node near coord/.append style={anchor=\anchor,text=black,font=\small},
                visualization depends on={value \thisrow{anchor} \as \anchor},
                only marks,
                point meta=explicit symbolic
            ] table[meta=lang,x index=1,y index=0] {
                loss logsize lang anchor
                1.9260400616332756 32.43295940727611 ar east
                3.42399222923749 33.21723071622067 {} east
                3.490658800393324 34.169925001442316 {} west
                2.4372046754538624 32.88752527074159 {} south
                2.5844004656577452 36.357552004618086 en north
                2.5728155339805832 33.807354922057606 es west
                4.269771955361466 33.807354922057606 {} north
                3.94668060637742 31.32192809488736 hi north
                2.6019514635976976 35.523561956057016 ru north
                3.4359538207806475 28.375039431346924 sw south
                3.446515190196585 33.12101540096137 th east
                4.3654822335025406 32.43295940727611 tr south
                4.30107526881721 29.78790255939143 ur north
                2.196704942586123 34.807354922057606 {} south
                2.1313941825476433 33.807354922057606 zh north
            };
            \addlegendentry{Grad (shared)}
            \addplot [
                lyyblue,thick,mark=square*,
                nodes near coords,
                scatter,
                every node near coord/.append style={anchor=\anchor,text=black,font=\small},
                visualization depends on={value \thisrow{anchor} \as \anchor},
                only marks,
                point meta=explicit symbolic,
            ] table[meta=lang,x index=1,y index=0] {
                loss logsize lang anchor
                2.568053415511039 32.43295940727611 {} south
                4.249635745507519 33.21723071622067 bg south
                3.6381514257620456 34.169925001442316 de {south west}
                2.4123352399900468 32.88752527074159 el north
                2.6775320139697384 36.357552004618086 en south
                2.791262135922323 33.807354922057606 {} west
                4.245511887433289 33.807354922057606 fr {south west}
                4.6785154208050114 31.32192809488736 hi south
                3.2524393294971254 35.523561956057016 ru {south west}
                8.383727322704776 28.375039431346924 sw south
                3.54863415879499 33.12101540096137 {} south
                4.086294416243664 32.43295940727611 {} north
                6.231044940722371 29.78790255939143 ur north
                1.9970044932601207 34.807354922057606 vi north
                3.159478435305917 33.807354922057606 zh west
            };
            \addlegendentry{$L_0$ (non-shared)}
        \end{axis}
    \end{tikzpicture}
    \caption{Accuracy loss on each language of XNLI vs. the logarithm of their pre-training corpus sizes.}
    \label{fig:size}
\end{figure}

\paragraph{Our improved $L_0$ regularization-based pruning can further boost the performance.}
In Section~\ref{sub-sec:improved-l0}, we propose an improved $L_0$ regularization to solve the drawbacks of standard $L_0$. Table \ref{tab:l0} shows the results.
Through the sparsity constraint, we can control the model sparsity to be the desired value $t=50\%$ instead of 20\% (the closest we could have using vanilla $L_0$).
And along with diverse subnetwork, the improved $L_0$ can even consistently improve the fine-tuning results.
Appendix \ref{app:diverse} visualizes how subnetworks differ between two languages after applying the diversity loss term. 
Moreover, integrating with distillation (the last row of Table~\ref{tab:l0}) can further improve the results.

\subsection{Analysis}
\label{sub-sec:analysis1}

\paragraph{Why does regularization-based pruning perform poorly?}
Since regularization-based pruning learns the subnetwork from scratch, we believe its poor performance results from the low-resource languages.
We choose XNLI with the translate-train-all setting for empirical verification.
On the one hand, the translate-train-all setting ensures that each language has the same dataset for fine-tuning (except for NER and POS).
This way eliminates the difference in fine-tuning.
On the other hand, among all tasks except NER and POS, XNLI covers more languages.

Figure \ref{fig:size} supports our hypothesis.
It shows the accuracy loss and corpus size of each language in regularization-based and gradient-based pruning.
We observe that for regularization-based pruning accuracy loss strongly correlates with pre-training dataset size (a value of 0.83 for Pearson's $\tau$), while it is not for gradient-based pruning.



\begin{figure}[t!]
    \centering
    \begin{tikzpicture}
        \begin{axis}[
            width=0.33\textwidth,height=0.305\textwidth,
            legend cell align={left},
            legend pos=outer north east,
            enlargelimits=0.1,
            legend style={
                font=\small,
                draw=none,
                column sep=3pt,
                legend columns=1,
                cells={align=left},
            },
            yticklabel style={/pgf/number format/fixed,/pgf/number format/precision=1},
            ylabel={Sparsity [\%]},
            ylabel near ticks,
            xlabel={Layer},
            xlabel near ticks,
            xmajorgrids=true,
            ymajorgrids=true,
            grid style=dashed,
            xtick={2,4,...,12},
            xmin=1,xmax=12,
            every tick label/.append style={font=\small},
            label style={font=\small},
            ylabel style={yshift=0pt},
        ]
            \addplot [lyyred,thick,mark=*] coordinates {
                (1,0.0)
                (2,0.0)
                (3,0.0)
                (4,0.0)
                (5,0.0)
                (6,0.0)
                (7,0.0)
                (8,0.0)
                (9,0.0)
                (10,0.0)
                (11,0.0)
                (12,0.0)
            };\addlegendentry{Grad (shar-\\ed) Head}
            \addplot [lyyred,thick,mark=square] coordinates {
                (1,100-73.58333587646484)
                (2,100-93.41666412353516)
                (3,100-87.33333587646484)
                (4,100-70.0)
                (5,100-53.583335876464844)
                (6,100-41.916664123535156)
                (7,100-45.08333206176758)
                (8,100-60.91666793823242)
                (9,100-59.41666793823242)
                (10,100-86.0)
                (11,100-74.83333587646484)
                (12,100-71.66666412353516)
            };\addlegendentry{$L_0$ (non-\\shared)\\ Head}
            \addplot [lyyblue,thick,densely dashed,mark=*,mark options={solid}] coordinates {
                (1,100-14.485676765441895)
                (2,100-24.674480438232422)
                (3,100-43.001304626464844)
                (4,100-38.216148376464844)
                (5,100-28.74349021911621)
                (6,100-21.126300811767578)
                (7,100-14.876301765441895)
                (8,100-13.28125)
                (9,100-15.071614265441895)
                (10,100-21.321613311767578)
                (11,100-18.912761688232422)
                (12,100-40.0390625)
            };\addlegendentry{Grad (shar-\\ed) Hidden}
            \addplot [lyyblue,thick,densely dashed,mark=square,mark options={solid}] coordinates {
                (1,100-41.51692581176758)
                (2,100-46.80338668823242)
                (3,100-56.37044143676758)
                (4,100-49.43684768676758)
                (5,100-38.65071487426758)
                (6,100-37.5380859375)
                (7,100-38.1513671875)
                (8,100-38.469078063964844)
                (9,100-43.482093811035156)
                (10,100-50.916015625)
                (11,100-48.621742248535156)
                (12,100-53.8076171875)
            };\addlegendentry{$L_0$ (non-\\shared)\\ Hidden}
        \end{axis}
    \end{tikzpicture}
    \caption{Sparsity of each layer pruned by two pruning algorithms.}
    \label{fig:compare}
\end{figure}
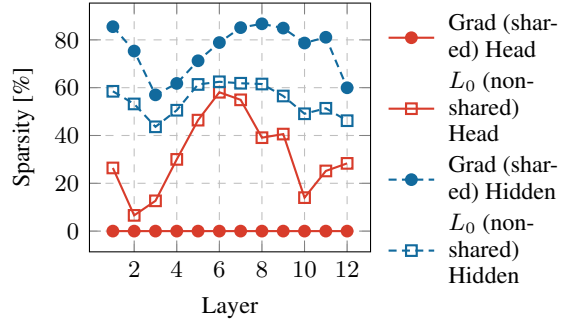

\paragraph{Where does pruning methods behave differently?}
In Figure \ref{fig:compare}, we compare in which aspect different pruning algorithms behave differently.
Figure \ref{fig:compare} shows the sparsity of each component (attention heads and hidden units) at each layer. Interestingly, we see that gradient-based pruning preserves all attention heads and only a tiny number of hidden units, while regularization-based pruning prunes heads and hidden units more evenly.
Though previous works \cite{michel2019sixteen,voita-etal-2019-analyzing} have suggested that most attention heads have little impact on the final performance of monolingual models, our results show that this is not the case for XLM-R.
Besides, both pruning methods tend to drop more in the middle layers.

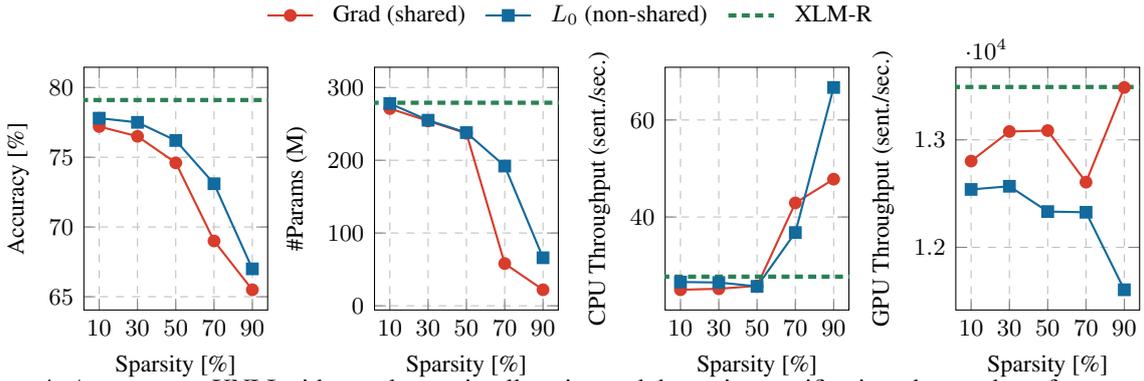
\begin{figure*}[t!]
    \centering
    \makeatletter
    \let\ref\@refstar
    \ref{grouplegend}
    \makeatother
    \vspace{-0.2in}
    \begin{tikzpicture}
        \begin{groupplot}[
            group style={group size=4 by 1, horizontal sep=40pt},
            width=1.0\textwidth,
            height=0.3\textwidth,
            legend cell align={left},
            legend pos=north west,
            enlargelimits=0.1,
            legend style={
                font=\small,
                draw=none,
                column sep=5pt,
                legend columns=3,
            },
        ]
        \nextgroupplot[
            width=0.25\textwidth,height=0.3\textwidth,
            yticklabel style={/pgf/number format/fixed,/pgf/number format/precision=1},
            ylabel={Accuracy [\%]},
            ylabel near ticks,
            xlabel={Sparsity [\%]},
            xlabel near ticks,
            xmajorgrids=true,
            ymajorgrids=true,
            legend pos=south west,
            grid style=dashed,
            xtick=data,
            every tick label/.append style={font=\small},
            label style={font=\small},
            ylabel style={yshift=0pt},
            legend to name=grouplegend,
            xmin=10,xmax=90,
            ymax=80,
        ]
            \addplot [lyyred,thick,mark=*] coordinates {
                (10,77.2) (30,76.5) (50,74.6) (70,69.0) (90,65.5)
            };\addlegendentry{Grad (shared)}
            \addplot [lyyblue,thick,mark=square*] coordinates {
                (10,77.8) (30,77.5) (50,76.2) (70,73.1) (90,67.0)
            };\addlegendentry{$L_0$ (non-shared)}
            \addplot [lyygreen,densely dashed,ultra thick,mark=none,samples=2,domain=0:100,y domain=65:80] {79.1};\addlegendentry{XLM-R}
        \nextgroupplot[
            width=0.25\textwidth,height=0.3\textwidth,
            yticklabel style={/pgf/number format/fixed,/pgf/number format/precision=1},
            ylabel={\#Params (M)},
            ylabel near ticks,
            xlabel={Sparsity [\%]},
            xlabel near ticks,
            xmajorgrids=true,
            ymajorgrids=true,
            grid style=dashed,
            xtick=data,
            every tick label/.append style={font=\small},
            label style={font=\small},
            ylabel style={yshift=0pt},
            xmin=10,xmax=90,
            ymax=300,
        ]
            \addplot [lyyred,thick,mark=*] coordinates {
                (10,271) (30,254) (50,237) (70,58) (90,22)
            };
            \addplot [lyyblue,thick,mark=square*] coordinates {
                (10,278) (30,255) (50,238) (70,192) (90,66)
            };
            \addplot [lyygreen,densely dashed,ultra thick,mark=none,samples=2,domain=0:100] {279};
        \nextgroupplot[
            width=0.25\textwidth,height=0.3\textwidth,
            yticklabel style={/pgf/number format/fixed,/pgf/number format/precision=1},
            ylabel={CPU Throughput (sent./sec.)},
            ylabel near ticks,
            xlabel={Sparsity [\%]},
            xlabel near ticks,
            xmajorgrids=true,
            ymajorgrids=true,
            grid style=dashed,
            xtick=data,
            every tick label/.append style={font=\small},
            label style={font=\small},
            ylabel style={yshift=0pt},
            xmin=10,xmax=90,
        ]
            \addplot [lyyred,thick,mark=*] coordinates {
                (10,25.0) (30,25.2) (50,25.8) (70,42.9) (90,47.8)
            };
            \addplot [lyyblue,thick,mark=square*] coordinates {
                (10,26.6) (30,26.5) (50,25.7) (70,36.8) (90,66.7)
            };
            \addplot [lyygreen,densely dashed,ultra thick,mark=none,samples=2,domain=0:100] {27.7};
        \nextgroupplot[
            width=0.25\textwidth,height=0.3\textwidth,
            yticklabel style={/pgf/number format/fixed,/pgf/number format/precision=1},
            ylabel={GPU Throughput (sent./sec.)},
            ylabel near ticks,
            xlabel={Sparsity [\%]},
            xlabel near ticks,
            xmajorgrids=true,
            ymajorgrids=true,
            grid style=dashed,
            xtick=data,
            every tick label/.append style={font=\small},
            label style={font=\small},
            ylabel style={yshift=0pt},
            xmin=10,xmax=90,
        ]
            \addplot [lyyred,thick,mark=*] coordinates {
                (10,12802.4) (30,13077.7) (50,13086.1) (70,12606.2) (90,13487.2)
            };
            \addplot [lyyblue,thick,mark=square*] coordinates {
                (10,12538.8) (30,12568.2) (50,12332.8) (70,12326.3) (90,11605.3)
            };
            \addplot [lyygreen,densely dashed,ultra thick,mark=none,samples=2,domain=0:100] {13491.7};
        \end{groupplot}
    \end{tikzpicture}
    \caption{Accuracy on XNLI with translate-train-all setting and dynamic sparsification, the number of parameters (\#Params), CPU and GPU throughput (the number of sentences per second) vs. the sparsity.}
    \label{fig:sparsity}
\end{figure*}

\section{Toward Efficient Pruning}
\label{sec:dyna}

\subsection{Dynamic Sparsification}
\label{sub-sec:dyna}

In practice, we may need models with different sparsities to fit various resource constraints or compare a set of methods.
Nevertheless, existing pruning techniques must train the model independently for each sparsity level, which is prohibitive for large models. Here we propose \emph{Dynamic Sparsification} (\textbf{DS} for short), a method that trains the model once but allows inference with any level of sparsity.

Section \ref{sec:pruning} shows that both gradient-based and regularization-based pruning follow the same procedure: we first determine a threshold, then get the importance score for each component, and set the gating variable to 1 if its score is larger than that threshold and 0 otherwise.
By adjusting the threshold, one can obtain networks with any sparsity.

Based on this, we model a gating variable $g$ as: 
\begin{equation}
    g = f(\alpha + t\theta)
    \label{eqn:gate}
\end{equation}
where $\alpha$ is a trainable importance score as in regularization-based pruning, $t$ is the targeted network size (which is one minus the sparsity), $t\theta$ is the threshold with a learnable $\theta$, $f$ is a function with output ranging between 0 and 1.
We choose $f$ to be Eqs. \ref{eqn:s} - \ref{eqn:g} because it enables us to optimize $\alpha$ and $\theta$ via $L_0$ regularization.
If $\alpha$ and $\theta$ are set properly, Eq. \ref{eqn:gate} will automatically determine whether its corresponding component should be activated under the targeted network size $t$.

Then is how to find $\alpha$ and $\theta$ using pruning algorithms.
We know that pruning algorithms could rank different components by their importance scores.
Based on this ranking, we identify the boundary network size that a specific component will be activated (denoted as $\hat{t}$) and will not.
These two conditions form a system of linear equations in two unknowns $\alpha$ and $\theta$:
\begin{equation}
\left\{\begin{aligned}
f\left(\alpha + \hat{t}\theta\right) &= 1\\
f\left(\alpha + \left(\hat{t} - \delta\right)\theta\right) &= 0
\end{aligned}\right.
\label{eqn:question}
\end{equation}
where $\delta$ is the network size that one component contributes to, $\hat{t}$ is the boundary network size where the corresponding gating variable $g$ should be 1 if $t>\hat{t}$ and 0 if $t<\hat{t}-\delta$.
$\hat{t}$ equals the ranking divided by the total number of components.
Eq. \ref{eqn:question} has a closed-form solution for $\alpha$ and $\theta$:\footnote{Eq. \ref{eqn:answer} has the numerical stability issue and weighs different components equally (See Appendix \ref{app:impl} for the solution).}
\begin{equation}
\left\{\begin{aligned}
\alpha &= \left(1 - \hat{t}/\delta\right)f^{-1}(1) + (\hat{t}/\delta) f^{-1}(0) \\
\theta &= \left(f^{-1}(1) - f^{-1}(0)\right)/\delta
\end{aligned}\right.
\label{eqn:answer}
\end{equation}

\begin{table}[t]
    \centering
    \setlength{\tabcolsep}{3pt}
    {\small
    \begin{tabular}{lccccc}
        \toprule
        \makecell[c]{\textbf{Methods}} & XNLI & POS & NER & TyDiQA & \textbf{\underline{Avg}} \\
        \midrule
        Grad (shared) & \textbf{76.8} & \textbf{88.4} & \textbf{88.0} & \textbf{69.5/54.6} & \textbf{78.8} \\
        \quad + DS & 74.6 & 87.6 & 87.1 & 64.0/48.3 & 76.4 \\
        \midrule
        $L_0$ (non-shared) & \textbf{76.3} & \textbf{87.9} & \textbf{86.8} & 67.8/52.5 & \textbf{77.8} \\
        \quad + DS & 76.2 & \textbf{87.9} & 86.7 & \textbf{67.9/52.4} & 77.7 \\
        \bottomrule
    \end{tabular}
    }
    \caption{The results of gradient-based and regularization-based pruning with or without dynamic sparsification (Sparsity=50\%).}
    \label{tab:dyna}
\end{table}

Before training, we use gradient-based pruning to initialize $\alpha$ and $\theta$ via Eq. \ref{eqn:answer}.
If only gradient-based pruning is adopted, $\alpha$ and $\theta$ are then clamped and only the retained network parameters will be updated, otherwise they can be jointly optimized via regularization-based pruning.
During training, we sample different $t$s to train different sized subnetworks.
At inference, $t$ is set to the targeted network size to prune the model.
If one wants to extend DS to \texttt{non-shared} setting, he can prune for each language once and compute a unique set of $\alpha$ and $\theta$ for each language.

\subsection{Main Results}
\label{sub-sec:result2}

Table \ref{tab:dyna} (\texttt{+ DS} rows) shows the 50\% sparsity results after applying DS to the two pruning algorithms under their best performing pruning settings (according to Table \ref{tab:main}).
Surprisingly, we observe that gradient-based pruning with \texttt{shared} setting suffers from a significant loss, while regularization-based pruning with \texttt{non-shared} setting has almost no loss.
This is because DS shares the weights between subnetworks of different sparsities hurts the model capacity, and \texttt{non-shared} setting enlarges the subnetwork capacity by untying weights of different languages.
Due to the expensive cost of training models without DS, we only test the impact of DS on 50\% sparsity, but we compare it with other systems with a smaller size (See Appendix \ref{app:compare}).
The leftmost part of Figure \ref{fig:sparsity} shows more on how the two pruning methods trade accuracy for efficiency under various sparsities.

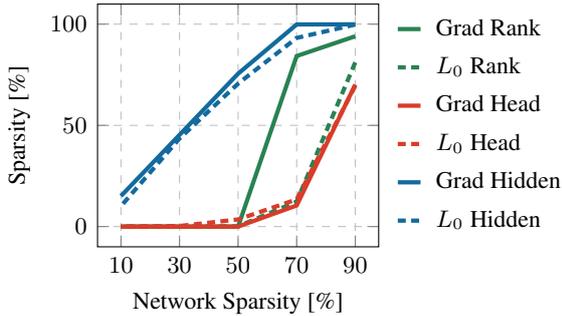
\begin{figure}[t!]
    \centering
    \begin{tikzpicture}
        \begin{axis}[
            width=0.33\textwidth,height=0.3\textwidth,
            legend cell align={left},
            enlargelimits=0.1,
            legend pos=outer north east,
            legend style={
                font=\small,
                draw=none,
                column sep=3pt,
                legend columns=1,
                minimum height=0.2in,
            },
            legend image post style={scale=0.5},
            yticklabel style={/pgf/number format/fixed,/pgf/number format/precision=1},
            ylabel={Sparsity [\%]},
            ylabel near ticks,
            xlabel={Network Sparsity [\%]},
            xlabel near ticks,
            xmajorgrids=true,
            ymajorgrids=true,
            grid style=dashed,
            xtick=data,
            every tick label/.append style={font=\small},
            label style={font=\small},
            ylabel style={yshift=0pt},
        ]
            \addplot [lyygreen,ultra thick] coordinates {
              (90,94.01041412353516) (70,84.24478912353516) (50,0.0) (30,0.0) (10,0.0)
            };
            \addlegendentry{Grad Rank}
            \addplot [lyygreen,densely dashed,ultra thick] coordinates {
              (90,81.11198425292969) (70,12.325515747070312) (50,0.11067962646484375) (30,0.0) (10,0.0)
            };
            \addlegendentry{$L_0$ Rank}
            \addplot [lyyred,ultra thick] coordinates {
              (90,70.13888549804688) (70,10.416671752929688) (50,0.0) (30,0.0) (10,0.0)
            };
            \addlegendentry{Grad Head}
            \addplot [lyyred,densely dashed,ultra thick] coordinates {
              (90,69.52777862548828) (70,13.416664123535156) (50,3.5277786254882812) (30,0.2916717529296875) (10,0.0)
            };
            \addlegendentry{$L_0$ Head}
            \addplot [lyyblue,ultra thick] coordinates {
              (90,99.91861724853516) (70,99.83995056152344) (50,75.52082824707031) (30,45.3125) (10,15.104171752929688)
            };
            \addlegendentry{Grad Hidden}
            \addplot [lyyblue,densely dashed,ultra thick] coordinates {
              (90,99.78255462646484) (70,93.19119262695312) (50,70.62676239013672) (30,43.51044464111328) (10,10.114311218261719)
            };
            \addlegendentry{$L_0$ Hidden}
        \end{axis}
    \end{tikzpicture}
    \caption{Sparsity of different components pruned by two pruning algorithms vs. the sparsity.}
    \label{fig:component}
\end{figure}

The second sub-figure from the left of Figure \ref{fig:sparsity} shows a non-linear relationship between the number of parameters and sparsity, as embeddings are not included in sparsity calculation \cite{prasanna-etal-2020-bert}.
Since embeddings are more important than most parts of the model and are very large (69\% of the overall parameters), the number of parameters remains high even when the encoder is quite sparse (Sparsity $\le 50\%$).
Pruning algorithms only start to prune these large embeddings when the encoder is very sparse (Sparsity $> 50\%$) and results in a great drop in the number of parameters, as shown in Figure \ref{fig:component}.

The two rightmost panels of Figure \ref{fig:sparsity} describe how the CPU and GPU throughput vary as the sparsity changes.
We observe a strong correlation between the CPU throughput and sparsity when the sparsity $\ge 50\%$.
However, there is no such trend observed when the sparsity $< 50\%$.
This might be due to the time consumption of irregular memory access out-weights the speed-up brought by the small tensor computation.

\begin{figure}[t!]
    \centering
    \begin{tikzpicture}
        \begin{axis}[
            width=0.33\textwidth,height=0.3\textwidth,
            legend cell align={left},
            enlargelimits=0.1,
            legend pos=outer north east,
            legend style={
                font=\small,
                draw=none,
                column sep=3pt,
                legend columns=1,
                minimum height=0.2in,     
            },
            yticklabel style={/pgf/number format/fixed,/pgf/number format/precision=1},
            ylabel={Layer Sparsity [\%]},
            ylabel near ticks,
            xlabel={Network Sparsity [\%]},
            xlabel near ticks,
            xmajorgrids=true,
            ymajorgrids=true,
            grid style=dashed,
            xtick=data,
            every tick label/.append style={font=\small},
            label style={font=\small},
            ylabel style={yshift=0pt},
            colormap/hot2,
            cycle list={[of colormap]},
            every axis plot/.append style={mark=none,ultra thick}, 
        ]
            \addplot coordinates {
                (90,80.19878387451172) (70,63.00998306274414) (50,54.009117126464844) (30,47.17708206176758) (10,20.928817749023438)
            };
            \addlegendentry{Layer 1}
            \addplot coordinates {
                (90,86.81532287597656) (70,59.361328125) (50,37.1015625) (30,26.254562377929688) (10,6.706382751464844)
            };
            \addlegendentry{Layer 3}
            \addplot coordinates {
                (90,83.90321350097656) (70,63.42404556274414) (50,44.292320251464844) (30,24.605247497558594) (10,6.0399322509765625)
            };
            \addlegendentry{Layer 5}
            \addplot coordinates {
                (90,90.67122650146484) (70,67.70616149902344) (50,54.71006774902344) (30,32.49284362792969) (10,5.916664123535156)
            };
            \addlegendentry{Layer 7}
            \addplot coordinates {
                (90,97.09809112548828) (70,71.06965637207031) (50,54.688148498535156) (30,30.35112762451172) (10,3.9839401245117188)
            };
            \addlegendentry{Layer 9}
            \addplot coordinates {
                (90,95.48025512695312) (70,70.349609375) (50,49.777774810791016) (30,21.191192626953125) (10,1.2979660034179688)
            };
            \addlegendentry{Layer 11}
        \end{axis}
    \end{tikzpicture}
    \caption{Sparsity of different layers pruned by regularization-based pruning vs. the sparsity.}
    \label{fig:layer}
\end{figure}
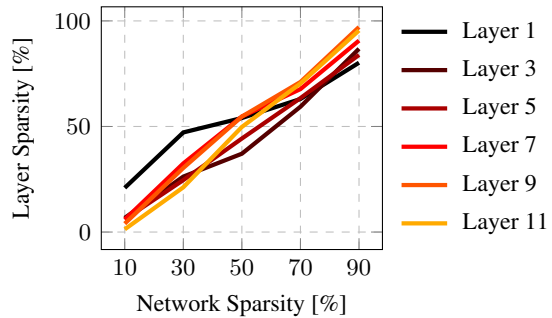

Interestingly, we see that sparse models show no acceleration on GPU even when the sparsity is high (e.g., 90\%).
Although pruning algorithms here optimize the model size instead of inference efficiency, it is expected that the resulting sparse models still have speedup as shown in CPU and in other work \cite{wang-etal-2020-structured}.
In Figure \ref{fig:layer}, we find that the highest sparsity of all layers is close to but not exactly 100\%.
This implies that \textbf{pruning tends to produce a deep and narrow model.}
Previous studies \cite{sanh2019distilbert,wang-etal-2020-hat,DBLP:conf/aaai/LiLXZ21} show that GPU throughput is more sensitive to the model height instead of its width.
This explains why we did not observe any acceleration even for a model with $1/10$ of the original size.

Though not shown in Table \ref{tab:main} and Figure \ref{fig:sparsity}, it is still possible to obtain actual speedup in GPU for sparse models.
\textbf{Previous observations on GPU throughput only hold for inference with the same batch size}.
In practice, the sparse models have a smaller memory footprint and we can use a larger batch size for higher parallelism.
For pruned models in Table \ref{tab:main}, a nearly 2$\times$ speedup is observed when we double the inference batch size.

In summary, Figure \ref{fig:sparsity} suggests that \textbf{the correlation between the model size and throughput is very week for XLM-R}: for model size, reducing the embedding size is important, but it has almost no impact on throughput (an $O(1)$ complexity table lookup); for throughput, compressing parts other than embeddings is more effective as shown in Figure \ref{fig:sparsity}, but they have much fewer parameters than the embeddings (193M parameters for embeddings vs. 86M for the others).
This advocates special care needed to be taken if one wants to compress and accelerate XLM-R simultaneously.

\subsection{Analysis}

Here we study what DS will prune under various sparsities.
Figure \ref{fig:component} shows which component (embeddings, attention heads and hidden units) will be preferred during pruning.
In general, gradient-based pruning behaves similar to regularization-based pruning: they first prune hidden units, and only prune attention heads and embeddings when the sparsity is high.
The main difference between them is that gradient-based pruning starts to prune embeddings earlier (at 70\% sparsity) than regularization-based pruning.
This explains why we observe a significant drop in performance for gradient-based pruning with 70\% sparsity (See the left of Figure \ref{fig:sparsity}): the model already lost much information at the beginning and there is no way to recover.

Figure \ref{fig:layer} shows how regularization-based pruning prunes each layer with DS.
Though we do not plot the curves of gradient-based pruning, its phenomenon is similar to regularization-basd pruning.
We find that regularization-based pruning behaves differently at low and high sparsity.
It first prunes bottom layers when the sparsity is low, then gradually shift to higher layers as the sparsity increases.
In the end, it retains more parameters in the bottom layers instead of the top layers.
This provides insight for future model design: \textbf{a pyramid structure is better when the model size is very small}.

\section{Conclusion}

In this work, we study three aspects of structured pruning on multilingual pre-trained models: settings, algorithms and efficiency.
Experiments show interesting phenomena: The best pruning setting depends on the choice of algorithms; The simplest pruning algorithm performs the best; A fast model does not mean it should be small.
We hope this work will give insight to future research.


\bibliography{anthology,custom}
\bibliographystyle{acl_natbib}

\appendix

\newpage

\section{Hyper-parameters}
\label{app:hyper}

\paragraph{Pre-training} We set $\lambda_1$ to 8 and $\lambda_2$ to 1 for $L_0$ regularization in 50\% sparsity.
If Dynamic Sparsification is applied, we set $\lambda_1$ to 128 and others remain the same.
The number of pre-training steps that tunes $\alpha$ only is 150K.

\paragraph{Fine-tuning} We perform a grid search to find the best hyper-parameter setting for each task (except for BUCC and Tatoeba, they do not need training).
We list the names of hyper-parameters as well as their search ranges below:
\begin{itemize}
    \item Learning rate: [1e-6, 2e-6, $\cdots$, 5e-5].
    \item Epoch: [5, 10] for cross-lingual transfer and 3 for translate-train-all.
\end{itemize}
We use a batch size of 32 for all experiments.

\section{Weighting $L_0$ Regularization}
\label{app:weight}

In practice, gating variables $g$ from different components should contribute differently to the overall $L_0$ regularization term $||G||_0$ in Eq. \ref{eqn:l0}, as they govern different weight matrices.
For example, disabling the head $i$ will remove $W^i_q,W^i_k,W^i_v$ and $W^i_o$, but disabling a hidden unit only eliminate a column of $W_1$ and a row of $W_2$.
So we weigh the regularization terms from attention heads by $64 \times 4$, $2$ for those from hidden units and $1$ for those from the embedding matrix.

\section{Language Family}
\label{app:family}

Table \ref{tab:family} is the language family information we used in Section \ref{sec:pruning}.
There are 15 different language families and one special \texttt{Missing} family in Table \ref{tab:family}.

\begin{table*}[t!]
    \centering
    {\small
    \begin{tabular}{ll|ll|ll}
        \toprule
        \multicolumn{1}{c}{Language} & \multicolumn{1}{c|}{Family} & \multicolumn{1}{c}{Language} & \multicolumn{1}{c|}{Family} & \multicolumn{1}{c}{Language} & \multicolumn{1}{c}{Family} \\
        \midrule
        af & Indo-European &
        am & Afro-Asiatic &
        ar & Afro-Asiatic \\
        as & Indo-European &
        az & Turkic &
        be & Indo-European \\
        bg & Indo-European &
        bn & Indo-European &
        bn-rom & Indo-European \\
        br & Indo-European &
        bs & Indo-European &
        ca & Indo-European \\
        cs & Indo-European &
        cy & Indo-European &
        da & Indo-European \\
        de & Indo-European &
        el & Indo-European &
        en & Indo-European \\
        eo & Constructed language &
        es & Indo-European &
        et & Uralic \\
        eu & Language isolate &
        fa & Missing &
        fi & Uralic \\
        fr & Indo-European &
        fy & Indo-European &
        ga & Indo-European \\
        gd & Indo-European &
        gl & Indo-European &
        gu & Indo-European \\
        ha & Afro-Asiatic &
        he & Afro-Asiatic &
        hi & Indo-European \\
        hi-rom & Indo-European &
        hr & Indo-European &
        hu & Uralic \\
        hy & Indo-European &
        id & Austronesian &
        is & Indo-European \\
        it & Indo-European &
        ja & Japonic &
        jv & Austronesian \\
        ka & Kartvelian &
        kk & Turkic &
        km & Austro-Asiatic \\
        kn & Dravidian &
        ko & Koreanic &
        ku & Indo-European \\
        ky & Turkic &
        la & Indo-European &
        lo & Kra-Dai \\
        lt & Indo-European &
        lv & Missing &
        mg & Missing \\
        mk & Indo-European &
        ml & Dravidian &
        mn & Missing \\
        mr & Indo-European &
        ms & Missing &
        my-zaw & Sino-Tibetan \\
        my & Sino-Tibetan &
        ne & Indo-European &
        nl & Indo-European \\
        no & Indo-European &
        om & Missing &
        or & Indo-European \\
        pa & Indo-European &
        pl & Indo-European &
        ps & Missing \\
        pt & Indo-European &
        ro & Indo-European &
        ru & Indo-European \\
        sa & Indo-European &
        sd & Indo-European &
        si & Indo-European \\
        sk & Indo-European &
        sl & Indo-European &
        so & Afro-Asiatic \\
        sq & Missing &
        sr & Indo-European &
        su & Austronesian \\
        sv & Indo-European &
        sw & Niger-Congo &
        ta & Dravidian \\
        ta-rom & Dravidian &
        te & Dravidian &
        te-rom & Dravidian \\
        th & Kra-Dai &
        tl & Austronesian &
        tr & Turkic \\
        ug & Turkic &
        uk & Indo-European &
        ur & Indo-European \\
        ur-rom & Indo-European &
        uz & Missing &
        vi & Austro-Asiatic \\
        xh & Niger-Congo &
        yi & Indo-European &
        zh-Hans & Sino-Tibetan \\
        zh-Hant & Sino-Tibetan & & & & \\
        \bottomrule
    \end{tabular}
    }
    \caption{The language family from \url{https://www.ethnologue.com/}. \texttt{Missing} means that there is no language family information of that language found in the website.}
    \label{tab:family}
\end{table*}

\section{Implementation of Dynamic Sparsification}
\label{app:impl}

Dynamic Sparsification described in Section \ref{sec:dyna} has two issues:
\begin{itemize}
    \item It assumes all components in the network contribute equally to the network size.
    But according to the discussion in Appendix \ref{app:weight}, different components relate to different numbers of weight matrices and each weight matrix has a different size.
    \item The solution of $\alpha$ and $\beta$ provided by Eq. \ref{eqn:answer} requires high precision in order to precisely activate just a single hidden unit by giving an appropriate sparsity.
    This fact brings difficulties in mixed-precision training as it easily causes the overflow issue.
\end{itemize}

Here we describe an improved version of Dynamic Sparsification for practical implementation.
The key difference between this improved version and the original one is the way it computes $\delta$ (the network size that a component should contribute to) and $\hat{t}$ (the network size where a component should be activated).

For $\delta$, we have:
\begin{enumerate}
    \item We associate a weight $w$ to each component, as done in Appendix \ref{app:weight}.
    \item Then $\delta=w/\left(\sum_{w' \in \bar{W}} w'\right)$, where $\bar{W}$ is the set of all $w$.
\end{enumerate}

\begin{figure*}[t!]
\centering
\includegraphics[scale=0.75]{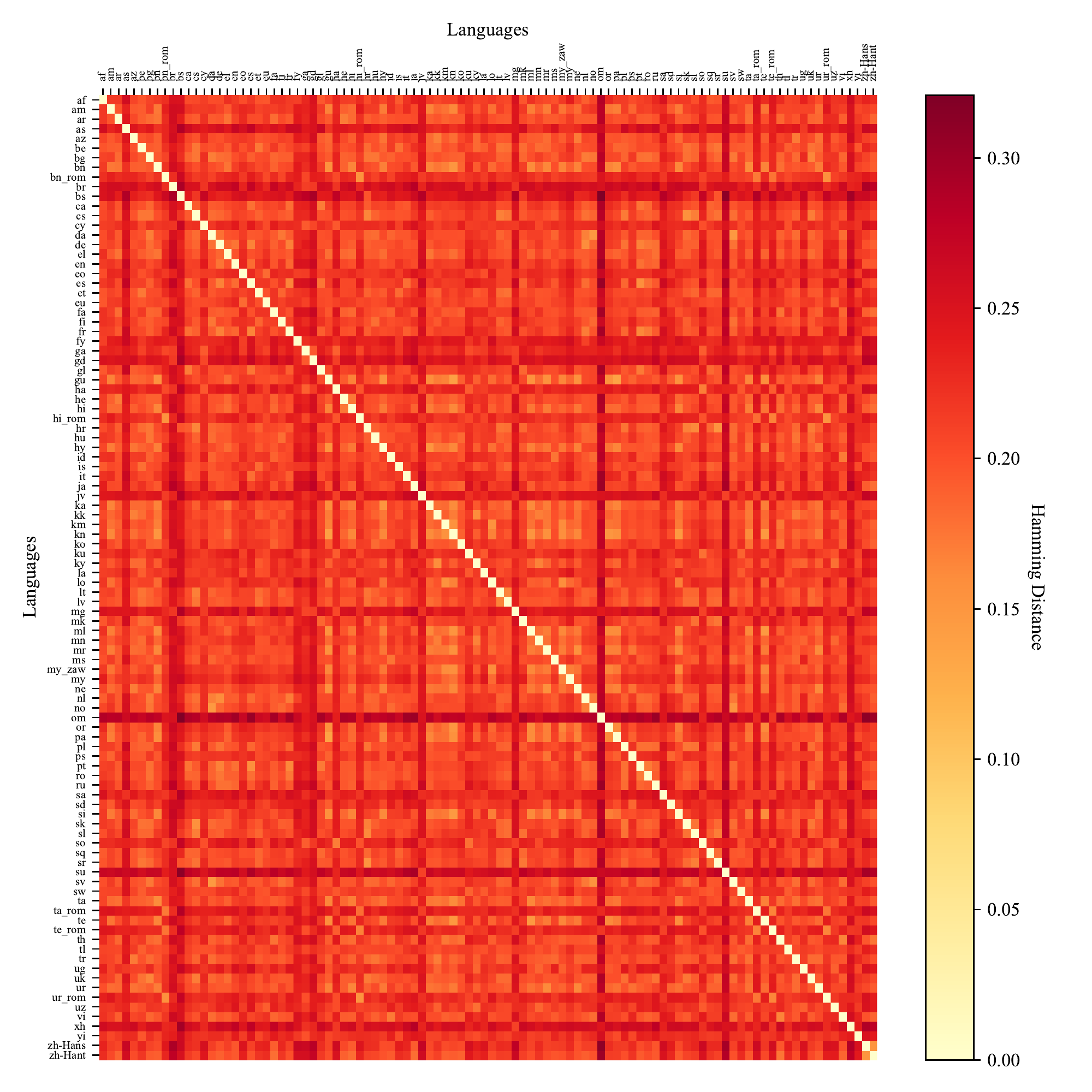}
\caption{Hamming distance between language subnetworks from regularization-based pruning with \texttt{non-shared} setting (Sparsity=50\%).}
\label{fig:sim}
\end{figure*}

\begin{table*}[t!]
    \centering
    \setlength{\tabcolsep}{2pt}
    {\small
    \begin{tabular}{lrcccccccccccccccc}
        \toprule
        \makecell[c]{System} & Sparsity & en & fr & es & de & el & bg & ru & tr & ar & vi & th & zh & hi & sw & ur & \underline{Avg} \\
        \midrule
        mMiniLMv1 & 70\% & \textbf{81.5} & 74.8 & 75.7 & 72.9 & 73.0 & 74.5 & 71.3 & 69.7 & 68.8 & 72.1 & 67.8 & 70.0 & 66.2 & 63.3 & 64.2 & 71.1 \\
        Grad (shared) + DS & 70\% & 69.0 & \textbf{76.0} & 71.9 & 73.0 & 70.8 & 70.3 & 70.8 & 70.0 & 68.4 & 66.7 & \textbf{71.0} & 66.7 & 68.1 & \textbf{65.4} & 64.1 & 62.4  \\
        $L_0$ (non-shared) + DS & 70\%  & 80.0 & 75.3 & \textbf{75.8} & \textbf{74.3} & \textbf{74.1} & \textbf{74.7} & \textbf{74.2} & \textbf{71.6} & \textbf{70.8} & \textbf{74.2} & 70.0 & \textbf{73.1} & \textbf{68.7} & 65.0 & \textbf{65.6} & \underline{\textbf{73.1}} \\
        \bottomrule
    \end{tabular}
    }
    \caption{XNLI results of mMiniLMv1, gradient-based (Grad) and regularization-based pruning ($L_0$) with Dynamic Sparsification (DS).}
    \label{tab:xnli}
\end{table*}

For $\hat{t}$, we have:
\begin{enumerate}
    \item We define a set of sparsities $\{s_0,s_2,\cdots,s_n\}$ (in sorted order) to be used at inference where $n$ is the number of all possible sparsities and $s_0=0$ and $s_n=1$, e.g., \{0\%, 10\%, $\cdots$, 100\%\}.
    \item A set of sparsity ranges can then be naturally derived from these sparsities, i.e., $\{s_0\sim s_1,\cdots,s_{i-1}\sim s_i,\cdots,s_{n-1}\sim s_n\}$.
    For example, given the set of sparsities \{0\%, 10\%, $\cdots$, 100\%\}, the set of ranges will be \{0\%$\sim$10\%, 10\%$\sim$20\%, $\cdots$, 90\%$\sim$100\%\}.
    \item For each sparsity range $s_{i-1}\sim s_i$, we find out all components that should be activated in that range, i.e., their original $\hat{t}$ must satisfy $s_{i-1}<\hat{t}\le s_i$ (considering their actual contributions to the total network size under the weighting scheme in Appendix \ref{app:weight}), and we denote these set of components as $C_i$.
    \item For all components $c \in C_i$, we assign their $\hat{t}=s_i$.
\end{enumerate}
The way we compute $\delta$ resolves the first issue by weighting the contribution to network size for each component.
And the way how $\hat{t}$ defined resolves the second issue by constraining the precision of sparsity and thus the precision of $\alpha$ and $\beta$.
Given $\hat{t}$ and $\delta$, we can use Eq. \ref{eqn:answer} to induce a solution that is numerical stable.

\section{Language Subnetwork Diversity}
\label{app:diverse}

Section \ref{sec:pruning} states that introducing a diversity loss term in Eq. \ref{eqn:diag} helps to diversify the subnetworks of each language.
To measure the distance between these subnetworks, we first choose the gating variables $G$ to represent a subnetwork.
We then calculate the Hamming distance between $G$s for each language pair.
Figure \ref{fig:sim} visualizes the results from the model pruned by our improved $L_0$ regularization.
We can see that subnetworks of different languages are indeed different.
Some languages are similar like \texttt{gu} and \texttt{bn}, but some are different like \texttt{bs} and \texttt{om}.
We also see that even for the most distant language pairs, they are still significantly overlapped (a Hamming distance around 0.3).
This indicates that sharing weights between languages is important.

\section{Comparison with Other Systems}
\label{app:compare}

Due to the expensive cost of pre-training models with different sparsities, we only compare the results with and without Dynamic Sparsification at 50\% sparsity, as shown in Table \ref{tab:dyna}.
Here in Table \ref{tab:xnli}, we compare our models trained by Dynamic Sparsification with mMiniLMv1\footnote{\url{https://github.com/microsoft/unilm/tree/master/minilm}} \cite{DBLP:conf/nips/WangW0B0020}, a system trained by advanced knowledge distillation techniques.
This mMiniLMv1 system has almost the same number of parameters as our 70\% sparsity models, and is also evaluated on XNLI.
Thus the comparison in Tables \ref{tab:xnli} and \ref{tab:main} helps to justify that Dynamic Sparsification does not degrade the performance much on different sparsity levels, especially for $L_0$ regularization with \texttt{non-shared} pruning setting.

\end{document}